\documentclass[10pt,twocolumn,letterpaper]{article}

\usepackage{iccv}              
\usepackage{graphicx}
\usepackage{amsmath}
\usepackage{amssymb}
\usepackage{booktabs}
\usepackage[linesnumbered, ruled,nofillcomment]{algorithm2e}
\usepackage{tabularx}
\usepackage{array}
\usepackage{multirow}
\usepackage{multicol}
\usepackage{xspace}
\usepackage{makecell}
\usepackage{float}
\usepackage{marvosym}

\makeatletter
\newcommand{\algorithmfootnote}[2][\footnotesize]{%
	\let\old@algocf@finish\@algocf@finish
	\def\@algocf@finish{\old@algocf@finish
		\leavevmode\rlap{\begin{minipage}{\linewidth}
				#1#2
		\end{minipage}}%
	}%
}
\makeatother
\newcommand\blfootnote[1]{%
	\begingroup
	\renewcommand\thefootnote{}\footnote{#1}%
	\addtocounter{footnote}{-1}%
	\endgroup
}

\usepackage [accsupp]{axessibility}
\definecolor{iccvblue}{rgb}{0.21,0.49,0.74}
\definecolor{darkgreen}{HTML}{009900}
\usepackage{enumitem}
\usepackage{bigstrut}
\usepackage{adjustbox}
\usepackage{indentfirst} 
\usepackage[pagebackref,breaklinks,colorlinks,allcolors=iccvblue]{hyperref}
\usepackage{xcolor}
\def\paperID{7212} 
\def\confName{ICCV}
\def\confYear{2025}

\title{CST Anti-UAV: A Thermal Infrared Benchmark for Tiny UAV Tracking in Complex Scenes}

\author{Bin Xie$^1$, Congxuan Zhang$^1$\textsuperscript{~\Letter}, Fagan Wang$^1$, Peng Liu$^2$, Feng Lu$^1$, Zhen Chen$^1$, Weiming Hu$^3$\\
	$^1$Nanchang Hangkong University, Nanchang, China \\
	$^2$Beihang University, Beijing, China \\
	$^3$Chinese Academy of Sciences, Beijing, China\\
{\tt\small zcxdsg@nchu.edu.cn}
}

\begin{document}
\twocolumn[{%
	\renewcommand\twocolumn[1][]{#1}%
	\maketitle 
	\begin{center} 
		\centering 
		\vspace{-3mm}
		\captionsetup{type=figure}
		\includegraphics[width=0.999\textwidth]{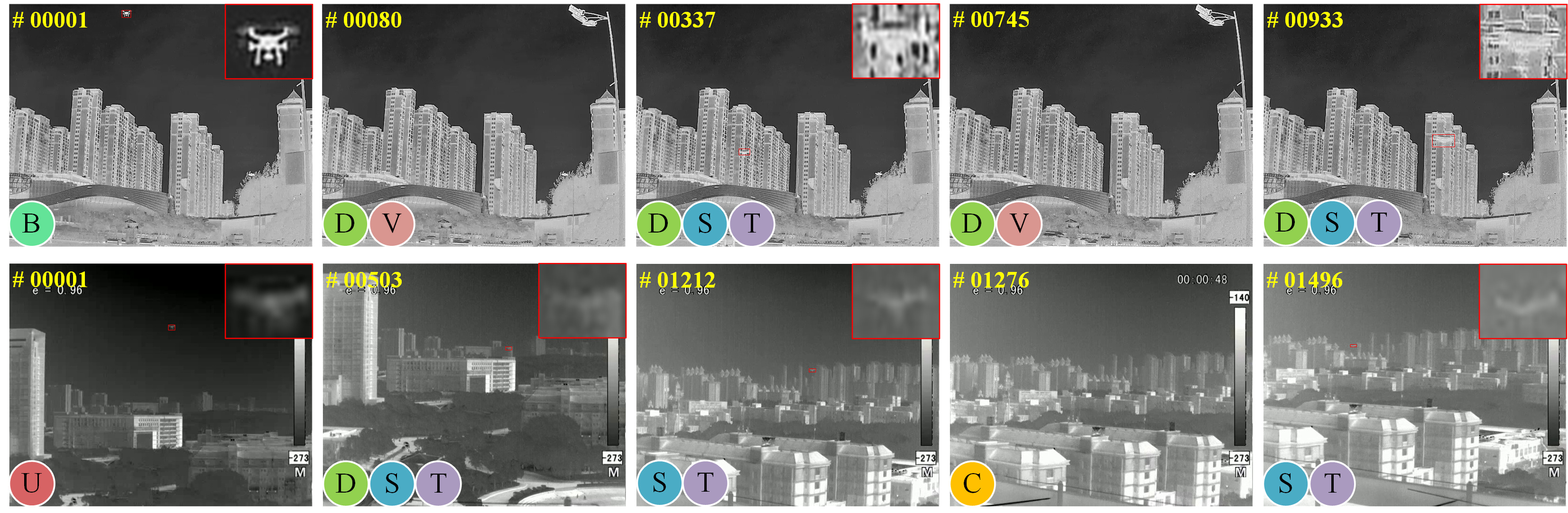} 
		\vspace{-7mm}
		\captionof{figure}{Examples of sequence clips from the CST Anti-UAV dataset. The UAV is annotated with the red bounding box. The most notable feature of CST Anti-UAV is complex scenes, as shown at the bottom, including the occlusion (C), complex dynamic background (D), scale variation (S), thermal crossover (T), and out-of-view (V). The background covers buildings (B), and urban areas (U). 
		}
		\vspace{3mm}
		\label{figure1}
	\end{center}
}]

\begin{abstract}
	\maketitle
	
	\blfootnote{\Letter~: Corresponding author.}
	The widespread application of Unmanned Aerial Vehicles (UAVs) has raised serious public safety and privacy concerns, making UAV perception crucial for anti-UAV tasks. However, existing UAV tracking datasets predominantly feature conspicuous objects and lack diversity in scene complexity and attribute representation, limiting their applicability to real-world scenarios. To overcome these limitations, we present the CST Anti-UAV, a new thermal infrared dataset specifically designed for Single Object Tracking (SOT) in \textbf{C}omplex \textbf{S}cenes with \textbf{T}iny UAVs (CST). It contains 220 video sequences with over 240k high-quality bounding box annotations, highlighting two key properties: a significant number of tiny-sized UAV targets and the diverse and complex scenes. To the best of our knowledge, CST Anti-UAV is the first dataset to incorporate complete manual frame-level attribute annotations, enabling precise evaluations under varied challenges. To conduct an in-depth performance analysis for CST Anti-UAV, we evaluate 20 existing SOT methods on the proposed dataset. Experimental results demonstrate that tracking tiny UAVs in complex environments remains a challenge, as the state-of-the-art method achieves only 35.92\% state accuracy, much lower than the 67.69\% observed on the Anti-UAV410 dataset. These findings underscore the limitations of existing benchmarks and the need for further advancements in UAV tracking research. The CST-Anti-UAV benchmark is now publicly available at \url{https://github.com/PCwenyue/CST-Anti-UAV}.
	\vspace{-3mm}
\end{abstract}


\begin{table*}[ht]
	\caption{Statistical comparison of CST Anti-UAV with current popular datasets for thermal infrared single object tracking in terms of the number of video sequences, bounding boxes, object size, and attributes. The number of tiny-sized objects in CST Anti-UAV is much larger than most thermal infrared tracking datasets. In the meantime, CST Anti-UAV is specifically for tracking UAVs in complex scenes and provides complete manual frame-level attribute annotations.}
	\centering
	\setlength{\arrayrulewidth}{0.8pt}
	\renewcommand{\arraystretch}{1.3}  
	\tabcolsep=3pt 
	\resizebox{\textwidth}{!}{%
		\centering
		\small
		\begin{tabular}{w{c}{3cm}||w{c}{1.8cm}|w{c}{1.8cm} |c|w{c}{2.5cm}|c|w{c}{2.5cm}|   cccc}
			\hline
			\multirow{2}{*}{Benchmark} & 
			\multirow{2}{*}{Sequences} & 
			\multirow{2}{*}{BBoxes} & 
			\multicolumn{2}{c|}{Sequence level attribute} & 
			\multicolumn{2}{c|}{Frame level attribute} & 
			\multicolumn{4}{c}{Object size} \\
			\cline{4-11}
			& & & Attribute & Sequences/Totals & Attribute & Frames/Totals & Tiny & Small & Normal & Large \\
			\hline
			PTB-TIR~\cite{ptb-tir}& 60 & 30k & 9 & 60/60 & - & - & 0 & 1883 & 6325 & 73925 \\
			LSOTB-TIR~\cite{lsotb-tir}& 1400 & 606k & 12 & 120/1400 & - & - & 0 & 7625 & 30910 & 568294 \\
			Anti-UAV~\cite{antiuav}& 318 & 293k & 7 & 318/318 & - & - & 858 & 6419 & 88181 & 198317 \\
			Anti-UAV410~\cite{antiuav410} & 410 & 438k & 6 & 410/410 & 6 & 701k/2628k & 17071 & 126649 & 97942 & 196735 \\
			\textbf{CST Anti-UAV}& \textbf{220} & \textbf{240k} & \textbf{6} & \textbf{220/220} & \textbf{6} & \textbf{1440k/1440k} & \textbf{78224} & \textbf{155202} & \textbf{11208} & \textbf{837} \\
			\hline
		\end{tabular}
		\vspace{-3mm}
		\label{table1}
	}
\end{table*}

\section{Introduction}
UAVs have rapidly proliferated due to their low cost and versatile applications in domains such as surveillance, delivery, and imaging~\cite{tuav,uav3,uav4,uav5}. However, incidents of unauthorized or malicious drone flights have infringed on privacy and endangered public safety, raising serious security and privacy concerns~\cite{uav2}. Consequently, there is a growing need for anti-UAV systems capable of real-time detection and tracking of unauthorized UAVs. UAV tracking is a crucial component of such systems, providing the location and trajectory of the target drone for interception or other countermeasures.

Most existing SOT datasets~\cite{otb50,otb100,alov,tc128} are based on visible light images, leading to unreliable tracking results in low-light or foggy conditions. Although some Thermal Infrared (TIR) datasets~\cite{vottir2016,vottir2015,pdt-atv,bu-tiv} have been proposed, they focus on general objects and may not be optimal for tiny UAV tracking. Anti-UAV~\cite{antiuav} is the first dataset specifically designed for UAV tracking. Anti-UAV410~\cite{antiuav410} was further expanded based on the Anti-UAV dataset by collecting some tiny UAVs and wild scenes. However, the existing datasets for tracking UAVs still have limitations as follows: 1) They lack a sufficient number of tiny UAV targets for effective training, thus limiting real-world applications. 2) The backgrounds of the existing datasets are relatively clean and quite simple, failing to capture the complexity of real-world tracking scenarios. 3) They have incomplete frame-level attribute annotations, hampering thorough evaluations of tracking methods.

To tackle the above problems, we propose a large-scale dataset for tracking UAVs in complex scenes, referred to as CST Anti-UAV.
We use popular TIR single-object tracking datasets, including PTB-TIR~\cite{ptb-tir}, LSOTB-TIR~\cite{lsotb-tir}, Anti-UAV~\cite{antiuav}, and Anti-UAV410~\cite{antiuav410}, as references to analyze the data statistics of the new CST Anti-UAV dataset. As shown in Table~\ref{table1}, the CST Anti-UAV dataset contains 220 sequences and over 240k high-quality bounding boxes, surpassing existing datasets in three aspects: 

\textbf{(1) A larger number of tiny-sized objects.} Similar to Anti-UAV410, we classify the object sizes into four categories based on the diagonal length of their bounding boxes: tiny (0,10], small (10,30], normal (30,50], and large (50,inf]. Our dataset contains 78,224 tiny objects, which is 4.5 times larger than existing large datasets. Adequate sample size guarantees robust training performance.

\textbf{(2) The diversity of complex tracking scenes.} In the real world, salient objects rarely appear while complex and dynamic scenes happen frequently. We first introduce the Complex Dynamic Background (CDB) attribute, which includes numerous dynamic distractors in the background. Furthermore, real-world tracking scenarios often involve multiple challenges. Existing datasets lack diversity in this aspect, driving us to collect sequences with complex challenge settings. As shown in Figure \ref{figure1}, we demonstrate examples of complex scenarios featuring multiple attributes and tiny-size objects.

\textbf{(3) The completeness of frame-level annotations.} We provide manual frame-level annotations for 6 attributes across all frames, resulting in a total of 1440k annotations. Unlike existing sequence-level annotations, frame-level annotations precisely reveal tracker responses to diverse challenges, offering more accurate guidance for future research. 

To analyze the proposed dataset, we retrained the 20 recent SOT methods on our dataset and conducted extensive evaluations. The experimental results demonstrate that complex tracking scenes and tiny-size objects make the SOT methods less pronounced. For example, SiamDT's~\cite{antiuav410} State Accuracy (SA) performance drops from 67.69\% on Anti-UAV410 to 35.84\% on CST Anti-UAV, while GlobalTrack~\cite{globaltrack} declines from 66.42\% to 35.92\%, consistently revealing the challenges brought by the CST Anti-UAV dataset. 

In summary, our main contributions are as follows:
\begin{itemize}
	\setlength\itemsep{0.1em}
	\item We present CST Anti-UAV, a large-scale thermal infrared dataset specifically tailored for UAV tracking. Characterized by tiny-sized objects and complex scenes, it aims to drive progress in anti-UAV research.
	\item To the best of our knowledge, CST Anti-UAV is the first dataset to provide complete manual frame-level attribute annotations, enabling fine-grained evaluations of existing methods under various challenging conditions.
	\item To conduct a comprehensive experimental analysis on CST Anti-UAV, we benchmarked a range of current tracking methods, providing important insights for future SOT direction.
\end{itemize}


\section{Related Work}

\noindent\textbf{Single object tracking methods.} Based on the latest developments in visual tracking, the tracking methods can be classified into three groups: Transformer-based methods~\cite{transt,hift,stark,sbt,atom,supersbt,swintrack,cswintt,simtrack} leverage the self-attention mechanism to model long-range dependencies and capture global context. Siamese methods~\cite{siamattn,stmtrack,ocean,alpha,siamfc++,siamfc,siamban,cgacd,siamrpnpp} employ twin networks to compare template and search features, balancing speed and accuracy. Other deep learning methods~\cite{kys,dimp,keeptrack,atom,eco,udt,mdnet,metasdnet} integrate diverse architectures like convolutional neural networks or attention mechanisms to address specific tracking challenges.

\noindent\textbf{Single object tracking benchmarks.} TrackingNet \cite{tracking} is a visible object tracking benchmark dataset based on YouTube-BoundingBoxes~\cite{ytbb}, containing over 30,000 video clips with 27 target categories. LaSOT~\cite{lasot} is a high-quality benchmark that includes 1,400 visible videos, with an average of around 2,500 frames per video. GOT10K~\cite{got10k} is a visible object tracking dataset that covers 563 types of objects, containing more than 1.5 million manually annotated bounding boxes. PTB-TIR~\cite{ptb-tir} is a thermal infrared pedestrian tracking dataset that contains 60 sequences collected from various devices. LSOTB-TIR~\cite{lsotb-tir} is currently the largest TIR object tracking dataset with 1,400 sequences, defining 12 challenge attributes.

\noindent\textbf{Benchmarks for anti-UAV.} Although several anti-UAV datasets have been proposed, they are still far from being practically applicable. Anti-UAV \cite{antiuav} is a dataset with visible and infrared dual-mode information, which consists of 318 fully labeled videos, making a significant contribution to maintaining airspace safety. However, most objects in the dataset are relatively large, and the backgrounds are clean and simple. DUT Anti-UAV \cite{dutantiuav} is divided into two tasks: detection and tracking. It contains 10,000 images for the detection task and 20 sequences for the tracking task. However, the small number of tracking sequences is insufficient for training tracking methods. Anti-UAV410 \cite{antiuav410} extends Anti-UAV to track drones in wild scenes. It has collected 150k drone objects and merged the infrared portion of the Anti-UAV dataset, ultimately containing 410 sequences with 438k objects. The collected data includes some tiny objects and wild scenes. Unfortunately, the number of tiny objects is too limited, and the scenes are predominantly in wilderness areas, lacking truly complex urban areas with dynamic distractors involving pedestrians and vehicles. The singular attribute of sequences fails to capture the complexity of real tracking scenes. Furthermore, the frame-level annotations are incomplete, hindering comprehensive evaluation of tracker performance across various attributes.

Given the limitations in existing datasets, we develop CST Anti-UAV, a comprehensive benchmark providing extensive tiny objects, complex real-world tracking scenarios, and full attribute annotations. The aim is to advance UAV tracking in complex scenes and bridge the gap toward real-world applications.

\begin{figure*}[ht] 
	\centering
	\includegraphics[width=0.9\textwidth]{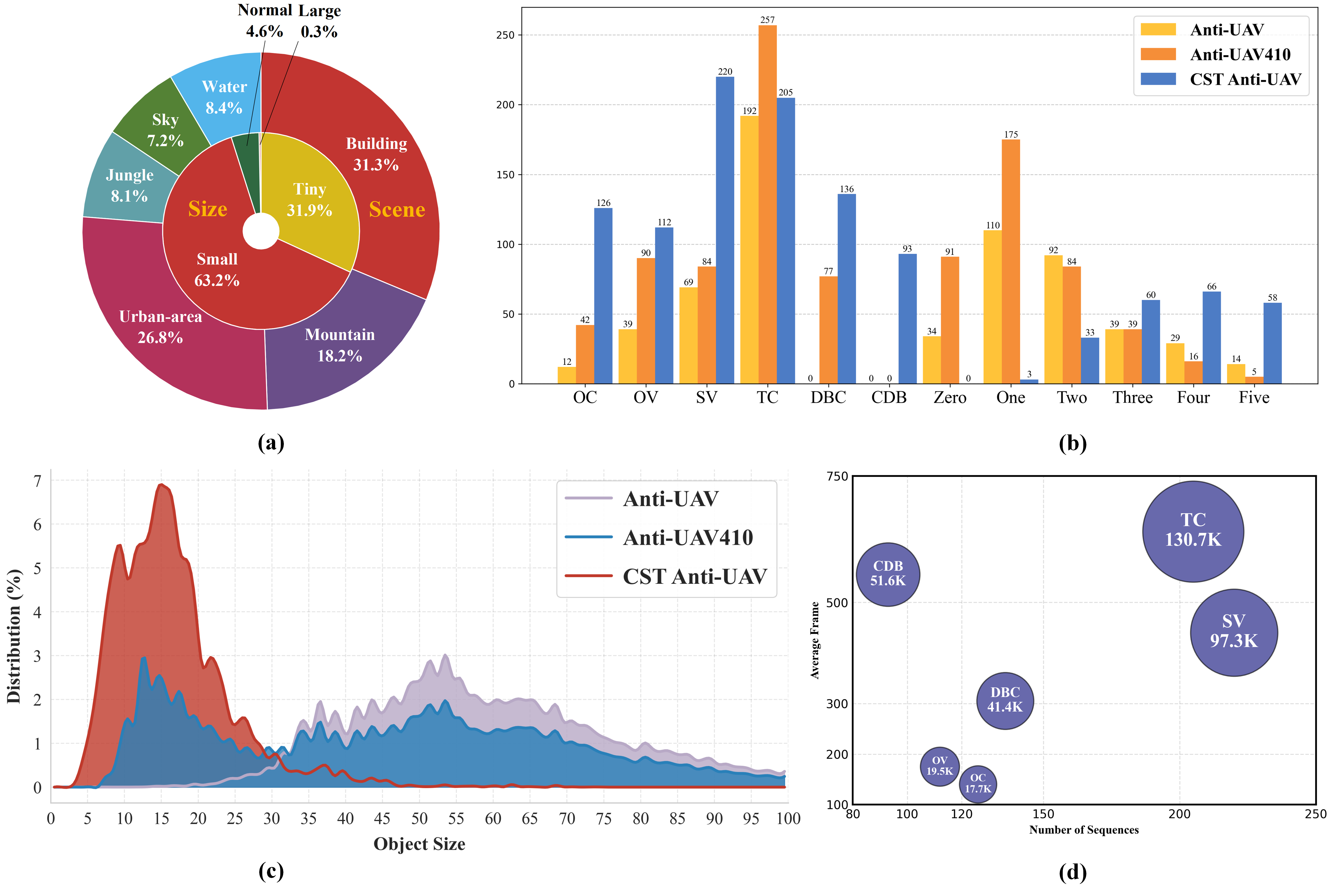} 
	\caption{Main features and statistics of the proposed dataset. (a) Distribution of scene (outer) and object size (inner). (b) A comparison of the sequence-level attributes of existing anti-UAV tracking datasets and our proposed dataset, emphasizing their significance and applicability for practical anti-UAV tasks in terms of video count. The numbers from Zero to Five on the table represent the number of attributes contained in a sequence. (c) Comparative analysis of the relative size distributions of objects in Anti-UAV, Anti-UAV410, and CST Anti-UAV. (d) Statistics of frame-level attributes in CST Anti-UAV. The area of each cycle denotes the number of total frames.} 
	\label{figure2}
	\vspace{-2mm} 
\end{figure*}


\section{CST Anti-UAV Benchmark}
\subsection{Data Acquisition and Analysis}

\noindent\textbf{Large-scale sequences with high diversity.} Existing datasets are~\cite{antiuav} typically recorded through a rotating platform equipped with a static camera, the acquisition scenes are limited only to a few selected. Therefore, we simultaneously adopt rotating platforms and professional drones equipped with infrared cameras (25 fps, 640 $\times$ 512) for data collection to ensure diversity. We carefully captured 220 sequences, comprising over 240k high-quality bounding boxes, and sequence lengths range from 600 to 2,062 frames (average 1,132 frames), covering both short-term and extended tracking scenarios. Furthermore, we classified UAV objects into four size categories, with their proportional distribution illustrated in the inner ring of Figure~\ref{figure2} (a). Notably, small and tiny objects are the domain, reflecting real-world tracking scenarios where UAVs are typically inconspicuous. Tiny objects are more difficult to track, while larger objects require fewer samples to achieve comparable performance. A key highlight is that our dataset contains 4.5 times more tiny objects than existing datasets. We further analyzed object size distribution compared to the Anti-UAV~\cite{antiuav} and Anti-UAV410~\cite{antiuav410} datasets, as shown in Figure~\ref{figure2} (c).

\noindent\textbf{Diverse and complex real-world scenarios.} Existing datasets typically feature simple and clean backgrounds, whereas our dataset captures rich and complex real-world scenes through a year-long data acquisition process. We comprehensively cover movement patterns including close-range, long-range, approaching, and receding trajectories, as well as diverse scenes such as urban areas, buildings, mountains, and skies, as quantified in the outer ring of Figure~\ref{figure2} (a). The dataset further incorporates seasonal variations (summer, autumn, winter), lighting conditions changes (day/night, sunny/cloudy), and extreme weather conditions like winds and fog. The background includes multiple dynamic distractors, such as moving pedestrians and vehicles that we introduced for the first time, birds, and airplanes. The complex backgrounds and significant weather or temperature variations in our dataset are crucial for training robust UAV tracking methods.

\begin{table}[ht]
	\centering
	\setlength{\arrayrulewidth}{0.8pt}
	\caption{Illustration of attribute annotation in CST Anti-UAV.}	
	\small
	\begin{tabular}{@{}>{\centering\arraybackslash}p{0.11\columnwidth}
			>{\raggedright\arraybackslash}p{0.80\columnwidth}@{}} 
		\hline
		\textbf{Attribute} & \textbf{Description} \\
		\hline 
		OC & \textbf{Occlusion:} the target is partially or fully occluded. \\
		OV & \textbf{Out-of-View:} the target leaves the view. \\
		SV & \textbf{Scale Variation:} the ratio of the bounding boxes of the first frame and the current frame is out of the range [0.66,1.5]. \\
		TC & \textbf{Thermal Crossover:} the target has a similar temperature to other objects or background surroundings. \\
		DBC & \textbf{Dynamic Background Clutter:} there are dynamic changes in the background around the target. \\
		CDB & \textbf{Complex Dynamic Background:} there are multiple dynamic non-target objects in the background. \\
		\hline 
	\end{tabular}
	\label{table2}
	\vspace{-3mm}
\end{table}

\noindent\textbf{Coarse-to-fine attribute annotation.} The challenges in our dataset can be summarized as six attributes, namely Occlusion (OC), Out-of-View (OV), Scale Variation (SV), Thermal Crossover (TC), Dynamic Background Clutter (DBC), and Complex Dynamic Background (CDB), the detailed meanings of each attribute are as shown in Table~\ref{table2}. In previous datasets, the backgrounds were mostly clean and simple. However, in real-world operational environments including aerial surveying and traffic monitoring, backgrounds inevitably contain complex dynamic objects like pedestrians and vehicles. We propose the novel concept of CDB which incorporates these dynamic objects. In comparison with existing anti-UAV datasets~\cite{antiuav,antiuav410} on sequence-level attributes, as shown in Figure~\ref{figure2} (b). The number of almost all attributes is significantly higher than in existing datasets. Notably, the CST Anti-UAV dataset contains 184 sequences with more than three attributes, compared to only 60 sequences in the Anti-UAV410 dataset. Furthermore, existing infrared tracking datasets only annotate attributes at the sequence level. In fact, only a few frames in a sequence may correspond to a specific challenge, roughly evaluating multiple different challenges across an entire sequence leads to flawed analysis results. We first introduce full frame-level attribute annotations. As shown in Figure~\ref{figure2} (d), we show the number of frames for each challenge attribute. 

\begin{figure}[htbp] 
	\centering 
	\includegraphics[width=\columnwidth]{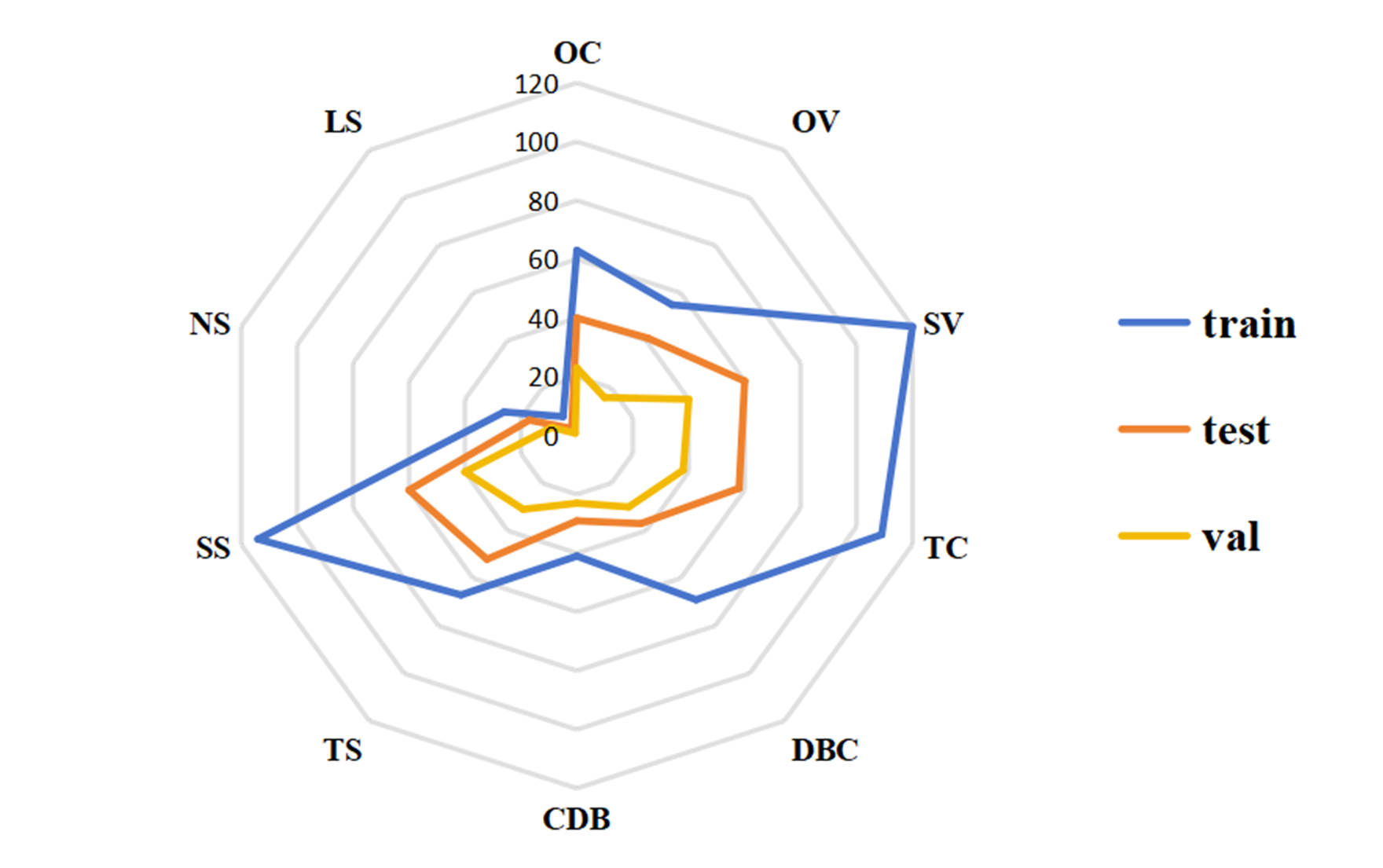} 
	\caption{The statistical distributions of the training set, validation set, and test set for each attribute are shown. The attribute distributions across the three subsets exhibit the same trends.} 
	\label{figure3} 
	\vspace{-3mm}
\end{figure}

\subsection{High-quality Human Annotation}
After collecting all the videos for the CST Anti-UAV dataset, we reviewed and edited them to reduce the number of inappropriate frames, resulting in over 240k frames. To ensure the quality of the annotations, we invested over 2,000 hours in the annotation process, which was divided into two tracks: 1) Bounding box annotation. Unlike existing datasets~\cite{antiuav,lsotb-tir} that rely on sparse annotation or algorithmic pre-labeling with manual correction, our annotation team manually obtained high-quality bounding boxes for each frame. Our verification team conducted continuous frame-by-frame checks until no issues remained. 2) Attribute annotation. To ensure consistency in evaluating the six proposed challenge attributes across frames, we assigned a single expert annotator to handle all attribute labeling, given that different annotators may exhibit varying tendencies. In total, we annotated 1440k attributes across all frames and provided 220 $\times$ 6 sequence-level attribute annotations.

\subsection{Dataset Splitting}
To enable fair comparisons in the evaluation of tracking methods and prevent overfitting, we split the CST Anti-UAV dataset into three sets: 120 sequences for training, 40 for validation, and 60 for testing, in accordance with the following rigorous criteria: 1) Each subset encompasses all scene categories and object size ranges. 2) Balanced distribution of challenges across the three sets. 3) The test set remains strictly independent, while training and validation sets derive from non-overlapping clips of shared sequences. The attribute distributions are illustrated in Figure~\ref{figure3}. The distribution trends of various attributes are consistent across all subsets, which ensures our training set can better capture the challenges of UAV tracking in real-world scenes.

\begin{table*}[ht]
	\caption{The performance of the state-of-the-art trackers on CST Anti-UAV and Anti-UAV410 datasets, evaluated by the metrics mSA, the Area Under the Curve (AUC) of Success Plots (S) and Precision Plots (P). All three metrics are reported in percentage form. 410 denotes the Anti-UAV410 dataset, while CST refers to the CST Anti-UAV dataset. The top three are highlighted in red, blue, and green, with bold text for emphasis. The asterisk (*) denotes an infrared method.}
	\centering
	\setlength{\arrayrulewidth}{0.8pt}
	\renewcommand{\arraystretch}{1.05}
	\small
	\tabcolsep=3pt 
	\begin{tabular}{w{c}{3cm}|w{c}{1.8cm}|w{c}{1.8cm}w{c}{1.8cm}w{c}{1.8cm}|w{c}{1.8cm}w{c}{1.8cm}w{c}{1.8cm}}
		\hline
		\multirow{2}{*}{Methods} & \multirow{2}{*}{Source} & \multicolumn{3}{c|}{Trained with 410 and test on 410} & \multicolumn{3}{c}{Trained with CST and test on CST} \\
		\cline{3-8}
		& & mSA & P(AUC) & S(AUC) & mSA & P(AUC) & S(AUC) \\
		\hline
		GlobalTrack~\cite{globaltrack}& AAAI20 & \textcolor{blue}{\textbf{66.42}} & \textcolor{blue}{\textbf{85.25}} & \textcolor{blue}{\textbf{65.11}} & \textcolor{red}{\textbf{35.92}} & \textcolor{red}{\textbf{58.72}} & \textcolor{red}{\textbf{35.38}} \\
		KYS~\cite{kys}                & ECCV20 & 45.12 & 61.31 & 44.30 & 26.28 & 40.20 & 25.93 \\
		PrDiMP~\cite{prdimp}          & CVPR20 & 59.31 & 76.77 & 59.09 & \textcolor{darkgreen}{\textbf{29.76}} & 42.55 & \textcolor{darkgreen}{\textbf{29.28}} \\
		Stark~\cite{stark}            & ICCV21 & 57.09 & 76.64 & 55.99 & 23.27 & 40.54 & 23.10 \\
		Stark-ST~\cite{stark}         & ICCV21 & 58.32 & 78.10 & 57.22 & 25.63 & 44.99 & 25.44 \\
		OSTrack-256~\cite{ostrack}    & ECCV22 & 52.47 & 70.39 & 51.44 & 26.34 & 46.95 & 26.16 \\
		OSTrack-384~\cite{ostrack}    & ECCV22 & 57.41 & 76.73 & 56.29 & 25.25 & 45.99 & 25.10 \\
		AiATrack~\cite{aiatrack}      & ECCV22 & 58.31 & 78.15 & 57.19 & 26.83 & 47.65 & 26.64 \\
		ToMP~\cite{tomp}              & CVPR22 & 56.44 & 73.70 & 55.28 & 22.84 & 39.81 & 22.66 \\
		Mixformer~\cite{mixformer}    & CVPR22 & 58.49 & 78.07 & 57.35 & 26.46 & 47.14 & 26.19 \\
		Mixformer-V2~\cite{mixformerv2}& NIPS23 & \textcolor{darkgreen}{\textbf{61.68}} & \textcolor{darkgreen}{\textbf{81.41}} & \textcolor{darkgreen}{\textbf{60.50}} & 29.39 & \textcolor{darkgreen}{\textbf{52.37}} & 29.09 \\
		DropTrack~\cite{droptrack}    & CVPR23 & 61.00 & 81.11 & 59.81 & 26.89 & 49.08 & 26.65 \\
		ARTrack-256~\cite{artrack}   & CVPR23 & 48.14 & 63.75 & 47.17 & 25.14 & 45.39 & 24.96 \\
		ARTrack-384~\cite{artrack}    & CVPR23 & 52.63 & 69.51 & 51.54 & 25.19 & 46.56 & 25.04 \\
		SeqTrack~\cite{seqtrack}      & CVPR23 & 50.50 & 67.26 & 49.49 & 26.53 & 47.67 & 26.38 \\
		GRM~\cite{grm}                & CVPR23 & 50.79 & 68.36 & 49.80 & 26.96 & 48.70 & 26.80 \\
		ROMTrack~\cite{romtrack}      & ICCV23 & 55.08 & 72.62 & 53.98 & 27.43 & 48.20 & 27.23 \\
		HIPTrack~\cite{hiptrack}      & CVPR24 & 57.78 & 77.59 & 56.64 & 25.62 & 47.48 & 25.42 \\
		AQATrack~\cite{aqatrack}      & CVPR24 & 56.64 & 75.33 & 55.52 & 27.60 & 50.43 & 27.45 \\
		ARTrackV2~\cite{artrackv2}    & CVPR24 & 52.51 & 70.41 & 51.49 & 22.14 & 39.73 & 22.03 \\
		SiamDT*~\cite{antiuav410}     & PAMI24 & \textcolor{red}{\textbf{67.69}} & \textcolor{red}{\textbf{87.63}} & \textcolor{red}{\textbf{66.34}} & \textcolor{blue}{\textbf{35.84}} & \textcolor{blue}{\textbf{57.96}} & \textcolor{blue}{\textbf{35.28}} \\
		Refocus-TIR*~\cite{refocus}   & TNNLS24 & 58.24 & 76.31 & 57.09 & 26.51 & 47.49 & 26.26 \\
		\hline
	\end{tabular}
	\label{table3}
\end{table*}


\section{Experiments}
To thoroughly analyze the proposed CST Anti-UAV dataset, we retrained 20 of the latest methods on both CST Anti-UAV and Anti-UAV410, followed by comprehensive performance evaluations.
\subsection{Experimental Settings}
\noindent\textbf{Evaluation metrics.} We use three common evaluation metrics, namely precision plot, success plot, and state accuracy, to assess tracker performance through One-Pass Evaluation (OPE). The calculation methods for precision plots and success plots are the same as those in~\cite{otb50}. State accuracy consistent with the previous method~\cite{antiuav}, $\text{\emph{IoU}}_{t}$ represents the Intersection over Union (IoU) between the predicted bounding box and its corresponding ground-truth, while $\text{\emph{v}}_{t}$ denotes the visibility flags of the ground-truth annotations. The tracker's predicted $\text{\emph{p}}_{t}$ is used to measure the state accuracy. The average state accuracy of all video sequences is the mSA.
\begin{equation}
	\resizebox{0.9\columnwidth}{!}{$\displaystyle
		SA = \sum_{t} \frac{I O U_{t} \times \delta\left(v_{t}>0\right) + p_{t} \times \left(1 - \delta\left(v_{t}>0\right)\right)}{T}.
		$}
\end{equation}

\noindent\textbf{Baseline tracker.} We conducted comprehensive evaluations of 20  single object trackers on both CST Anti-UAV and Anti-UAV410~\cite{antiuav410} datasets. The transformer trackers include Stark~\cite{stark}, Stark-ST~\cite{stark}, OSTrack~\cite{ostrack}, AiATrack~\cite{aiatrack}, MixFormer~\cite{mixformer}, MixFormerV2~\cite{mixformerv2}, DropTrack~\cite{droptrack}, ARTrack~\cite{artrack}, SeqTrack~\cite{seqtrack}, GRM~\cite{grm}, ROMTrack~\cite{romtrack}, HIPTrack~\cite{hiptrack}, AQATrack~\cite{aqatrack} and ARTrackV2~\cite{artrackv2}. The other deep-learning trackers contain GlobalTrack~\cite{globaltrack}, KYS~\cite{kys}, PrDiMP~\cite{prdimp}, ToMP~\cite{tomp}, and Refocus-TIR~\cite{refocus}. For the Siamese trackers, we have only selected SiamDT~\cite{antiuav410}, which is designed for TIR UAV tracking tasks. We conducted all experiments on a server equipped with 8 NVIDIA A100-PCIE-40GB GPUs and used the implementations, pre-trained models, and default configurations provided by the official code to ensure a fair comparison among the trackers.

\subsection{Overall Performance}

Table~\ref{table3} presents the results on both Anti-UAV410 and CST Anti-UAV datasets, including state accuracy, and the Area Under the Curve (AUC) of success and precision plots. Notably, trackers that achieve superior performance on Anti-UAV410 exhibit significant degradation when evaluated on the proposed CST Anti-UAV dataset. The poor performance on CST Anti-UAV is not only due to complex backgrounds and tiny objects in static images but also because of object disappearance and reappearance across the temporal domain. Based on the evaluation, SiamDT~\cite{antiuav410} and GlobalTrack~\cite{globaltrack} secured the top two positions with an absolute advantage on both datasets. This indicates that trackers based on long-term tracking have the potential to achieve better performance, as they search in the whole image, allowing them to obtain the disappeared object's reappearing position.

\begin{figure}[ht] 
	\vspace{-1mm}
	\centering 	
	\includegraphics[width=0.85\columnwidth]{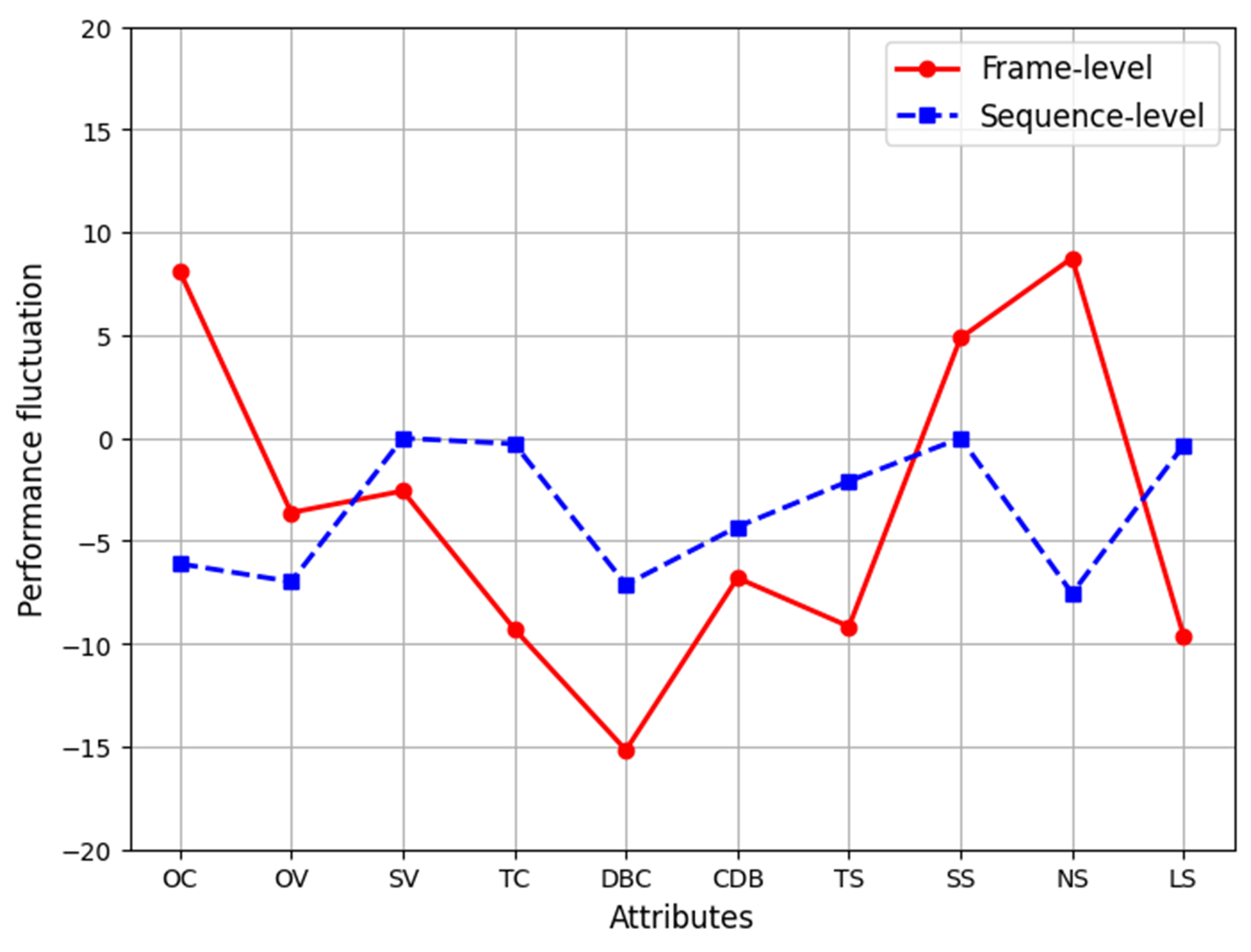} 
	\caption{Frame-level and sequence-level attribute SA results compared to overall SA results in terms of fluctuation.} 
	\label{figure4}
	\vspace{-4mm}
\end{figure}
In addition to experiments between different trackers, we further explored the impact of various tracker versions and different search region sizes on their performance. Stark-ST, which adds a dynamically changing template to capture temporal variations in targets, achieved an improvement in performance. This demonstrates that effectively utilizing temporal information can help address changes in object size and appearance. OSTrack-384~\cite{ostrack} and ARTrack-384~\cite{artrack} show notable performance improvements compared to OSTrack-256 and ARTrack-256 on Anti-UAV410 but have limited impact on CST Anti-UAV. We attribute this to the tiny size of the targets, which occupy fewer pixels in the image, making it harder for the network to effectively leverage its semantic representation capabilities.

\begin{table}[ht]
	\caption{When trained separately on the Anti-UAV410 and CST Anti-UAV datasets, the mSA(\%) performance of current trackers on the CST Anti-UAV test set. We selected ten of the methods for presentation. 410 denotes the Anti-UAV410 dataset, while CST refers to the CST Anti-UAV dataset.}
	\centering  
	\renewcommand{\arraystretch}{1.05}
	\setlength{\arrayrulewidth}{0.8pt}  
	\centering
	\resizebox{\columnwidth}{!}{%
		\small
		\begin{tabular}{w{c}{3cm}|c|c}
			\hline
			\multirow{2}{*}{Methods} & 
			\multicolumn{1}{c|}{Trained with 410} & 
			\multicolumn{1}{c}{Trained with CST} \\ 
			\cline{2-3}
			& Test on CST & Test on CST \\
			\hline
			GlobalTrack~\cite{globaltrack}&	30.96&	35.92\textcolor{red}{(+16.02\%)} \\
			OSTrack-256~\cite{ostrack}		&21.81	&26.34\textcolor{red}{(+20.77\%)}\\
			OSTrack-384~\cite{ostrack}&	21.29&	25.76\textcolor{red}{(+21.00\%)} \\
			SeqTrack~\cite{seqtrack}   &	20.36&	26.53\textcolor{red}{(+30.30\%)} \\
			GRM~\cite{grm}    &	21.64&	26.96\textcolor{red}{(+24.58\%)} \\
			ROMTrack~\cite{romtrack} &	21.13&	27.43\textcolor{red}{(+29.82\%)} \\
			AQATrack~\cite{aqatrack}  &	23.12&	27.60\textcolor{red}{(+19.38\%)} \\
			ARTrackV2~\cite{artrackv2} &19.14&22.14\textcolor{red}{(+15.67\%)} \\
			SiamDT~\cite{antiuav410}  &	30.95&	35.84\textcolor{red}{(+15.80\%)} \\
			Refocus-TIR~\cite{refocus}&	24.90&	26.51\textcolor{red}{(+6.47\%)} \\
			\hline
		\end{tabular}}
	\vspace{-3mm}
	\label{table4}
	
\end{table}

\begin{table*}[ht]
	\centering
	\caption{The performances mSA(\%) of baseline trackers on the CST Anti-UAV test set with different training datasets are evaluated across various frame-level attributes, including six attributes and four size categories. Details regarding the definitions of attributes can be found in Table~\ref{table2}. 410 denotes the Anti-UAV410 dataset, while CST refers to the CST Anti-UAV dataset. The top three are highlighted in red, blue, and green respectively, with bold text for emphasis. }
	\renewcommand{\arraystretch}{1.15}  
	\centering
	\setlength{\arrayrulewidth}{0.8pt}
	\resizebox{1\textwidth}{!}{%
		\small		
		\begin{tabular}{c|c|c|c|c|c|c|c|c|c|c|c|c}
			\hline
			\multirow{2}{*}{Methods} & \multirow{2}{*}{\parbox{1.2cm}{\centering Training\\dataset}} & \multicolumn{11}{c}{Frame-level attribute} \\ \cline{3-13} 
			&                                 & OC    & OV    & SV    & TC    & DBC   & CDB   & TS    & SS    & NS    & LS    & ALL   \\ \hline
			\multirow{2}{*}{KYS~\cite{kys}}     & 410                             & 29.45 & 54.55 & 21.49 & 13.56 & 6.92  & 12.77 & 23.81 & 29.43 & 9.25  & 20.26 & 26.28 \\ 
			& CST                             & 37.04 &  \textcolor{darkgreen}{\textbf{61.82}} & 21.68 & 13.75 & 7.23  & 10.80 &  \textcolor{blue}{\textbf{26.91}} & 29.17 & 13.83 &  \textcolor{darkgreen}{\textbf{19.67}} & 26.75 \\ \hline
			\multirow{2}{*}{AiATrack~\cite{aiatrack}} & 410                             & 30.37 & 8.52  & 20.40 & 14.32 & 6.56  & 14.56 & 13.26 & 26.78 & 42.74 & 11.27 & 22.37 \\  
			& CST                             & 37.42 & 16.64 & 25.79 & 18.49 &  \textcolor{darkgreen}{\textbf{12.15}} & 21.24 & 17.23 & 32.23 & 39.21 & 14.91 & 26.83 \\ \hline
			\multirow{2}{*}{ROMTrack~\cite{romtrack}} & 410                             & 25.63 & 8.99  & 16.85 & 12.81 & 7.68  & 17.69 & 11.58 & 25.08 & 31.05 & 14.01 & 21.13 \\ 
			& CST                             &  \textcolor{darkgreen}{\textbf{39.49}} & 18.66 & 25.78 & 17.69 & 10.33 & 19.69 & 16.79 & 32.32 & 33.69 & 14.66 & 27.43 \\ \hline
			\multirow{2}{*}{AQATrack~\cite{aqatrack}} & 410                             & 30.13 & 8.56  & 20.13 & 13.60 & 10.16 & 18.01 & 15.24 & 27.65 & 31.65 & 12.64 & 23.12 \\  
			& CST                             & 33.76 & 16.24 & 24.00 & 17.75 & 10.59 & 19.48 &  \textcolor{darkgreen}{\textbf{20.37}} & 32.07 & 26.82 & 14.65 & 27.60 \\ \hline
			\multirow{2}{*}{Mixformer-V2~\cite{mixformerv2}} & 410                          & 26.76 & 10.01 & 25.78 & 17.07 & 11.31 & 18.86 & 15.17 & 32.32 & 47.31 & 20.44 & 26.32 \\  
			& CST                             & 33.75 & 14.53 &  \textcolor{darkgreen}{\textbf{28.98}} &  \textcolor{darkgreen}{\textbf{22.10}} & 11.65 & 19.81 & 19.36 &  \textcolor{darkgreen}{\textbf{36.18}} &  \textcolor{darkgreen}{\textbf{41.48}} & 13.50 & 29.39 \\ \hline
			\multirow{2}{*}{PrDiMP~\cite{prdimp}}   & 410                             & 35.92 & 44.86 & 21.97 & 15.30 & 11.16 & 19.36 & 21.96 & 30.48 & 27.13 & 10.57 & 26.25 \\ 
			& CST                             & \textcolor{red}{\textbf{51.25}} & 57.26 & 27.86 & 16.99 & 11.50 &  \textcolor{darkgreen}{\textbf{21.81}} &  \textcolor{red}{\textbf{28.20}} & 33.45 & 29.85 & 12.94 & 29.76 \\ \hline
			\multirow{2}{*}{SiamDT~\cite{antiuav410}}   & 410                             & 33.07 & 60.43 & 33.60 & 22.12 & 14.62 & 24.23 & 12.89 & 39.32 & 59.11 & 41.25 & 30.95 \\ 
			& CST                             & 39.31 &  \textcolor{blue}{\textbf{64.54}} & \textcolor{red}{\textbf{39.24}} & \textcolor{red}{\textbf{27.22}} &  \textcolor{red}{\textbf{18.00}} & \textcolor{red}{\textbf{28.40}} & 14.98 &  \textcolor{red}{\textbf{46.51}} &  \textcolor{red}{\textbf{66.10}} &  \textcolor{red}{\textbf{38.34}} &  \textcolor{blue}{\textbf{35.84}} \\ \hline
			\multirow{2}{*}{GlobalTrack~\cite{globaltrack}} & 410                          & 32.86 & 57.71 & 31.36 & 20.38 & 12.77 & 23.67 & 13.81 & 39.12 & 59.73 & 20.64 & 30.96 \\  
			& CST                           &  \textcolor{blue}{\textbf{44.53}} & \textcolor{red}{\textbf{65.77}} &  \textcolor{blue}{\textbf{37.62}} &  \textcolor{blue}{\textbf{26.63}} &  \textcolor{blue}{\textbf{17.26}} &  \textcolor{blue}{\textbf{26.95}} & 15.83 &  \textcolor{blue}{\textbf{46.14}} &  \textcolor{blue}{\textbf{59.15}} &  \textcolor{blue}{\textbf{34.52}} &  \textcolor{red}{\textbf{35.92}} \\ \hline
	\end{tabular}}
	\label{table5}
	\vspace{-3mm}
\end{table*}

\subsection{Difficulty Level of CST Anti-UAV}
\label{sec:difficulty}

We conduct an experiment to further demonstrate the difficulty level of the CST Anti-UAV benchmark. We trained the 20 trackers on the existing most diverse anti-UAV tracking dataset, Anti-UAV410, and evaluated them on the CST Anti-UAV test set using the SA metric. The results of the ten selected methods are shown in Table \ref{table4}. For the results of all trackers, please refer to the supplementary materials. We observe a severe performance degradation when training on previous datasets. Existing datasets are inherently limited, lacking the diversity and complexity required to address the challenges posed by the CST Anti-UAV benchmark. We further quantified performance improvements through relative percentage increases. The results showed that 19 out of 22 trackers improved by over 10\%, with the highest growth reaching 30.3\%. These suggest that the current datasets are insufficient to represent tiny UAV perception and outline the need for complex, diverse datasets like CST Anti-UAV to advance the field.

\begin{figure*}[htbp] 
	\centering 
	\includegraphics[width=\textwidth]{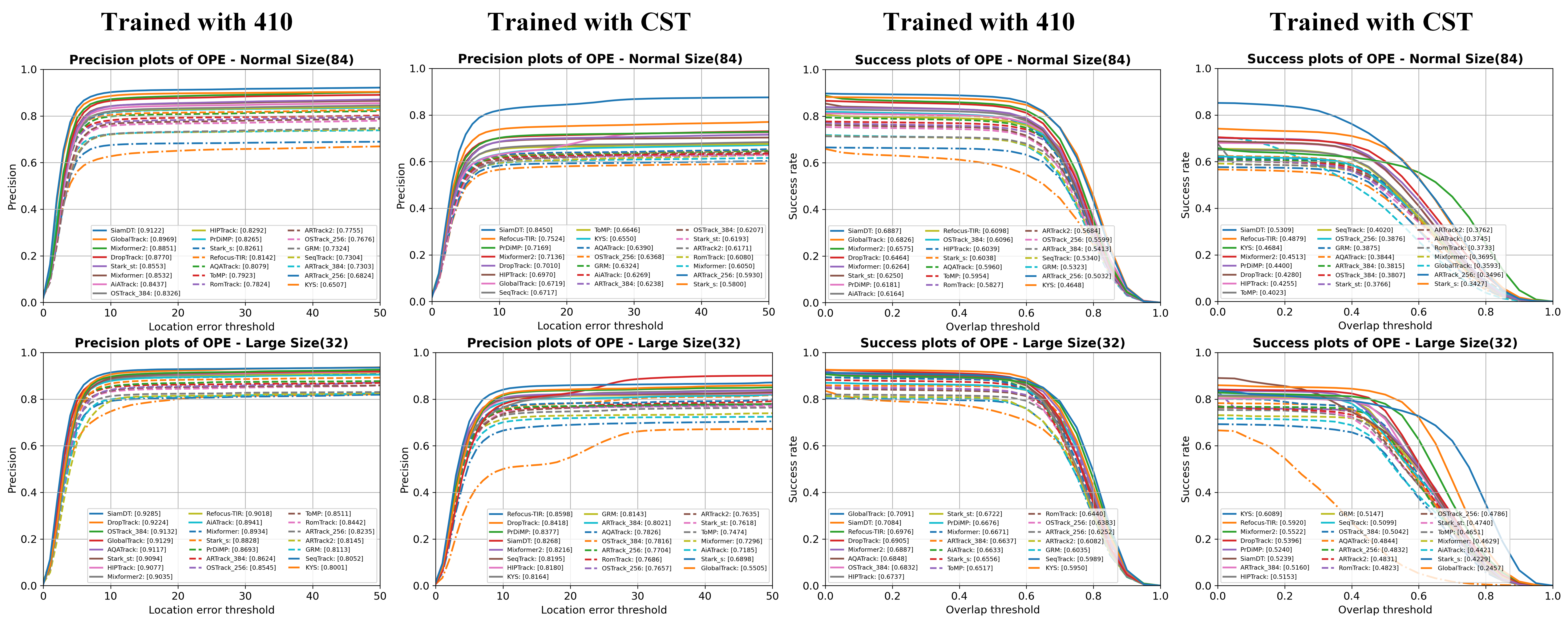} 
	\caption{Scale evaluation on Anti-UAV410 test set. Each column specifies the training dataset used.}
	\label{figure5} 
	\vspace{-3mm}
\end{figure*}

\subsection{Coarse-to-fine Attribute Evaluation}

To analyze the different challenges faced by trackers, all trackers were evaluated at both the sequence level and frame level based on six attributes and four object sizes on the CST Anti-UAV dataset, using state accuracy. We selected the results of the eight trackers for display, with Table~\ref{table5} showing the frame-level results. The sequence-level and complete frame-level results are available, please refer to them in the supplementary materials.

Experimental results show that sequence-level evaluations fluctuate little compared to the significant variations seen at the frame level which arises from that not all frames within a sequence contain challenging conditions. 
To illustrate the distinction between sequence-level and frame-level attribute evaluations, we quantified the deviations of both sequence-level and frame-level attribute-specific performance from the overall performance. As shown in Figure~\ref{figure4}, the frame-level exhibits analysis of more realistic variations. For attributes like scale variation and small size that span the entire test set, sequence-level analysis is equivalent to the overall performance and can even be misleading for occlusion and normal size. Thus, frame-level attribute annotations provide valuable guidance and actionable insights for improving tracker performance.

Based on the frame-level results, we observed a significant degradation in tracking performance for the attributes DBC, TC, CDB, OV, and tiny-sized objects. This indicates that tracking in complex scenarios and with tiny-sized objects remains a challenge for current trackers, demonstrating the effectiveness of the proposed CDB attribute. Among these, the OV attribute exhibits a pronounced bimodal distribution in performance, with peak performance at 65.77\% and minimum performance dropping to 14.53\%. This highlights the critical importance of re-localizing the target when it reappears. Occluded objects do not experience significant positional changes when reappearing, making them easier to locate than out-of-view ones. It is worth mentioning that with the help of the CST Anti-UAV training set, all evaluated trackers exhibit substantial performance improvements in complex scenarios and tiny object tracking. When addressing normal and large objects, most trackers achieve performance comparable to, or only marginally lower than that of trackers trained with Anti-UAV410.

Concretely, GlobalTrack~\cite{globaltrack} and SiamDT~\cite{antiuav410} still perform well across various attributes and object scales except for tiny objects, likely due to their global search modules. While this ensures strong target re-localization capability after the disappearance, the expanded search regions introduce more background noise,  leading to degraded performance on tiny-sized objects. KYS~\cite{kys} uses dense localized state vectors and a prediction module to propagate scene information across frames, enabling strong performance in tracking out-of-view attributes and tiny objects. PrDiMP~\cite{prdimp} combines probabilistic regression and DiMP~\cite{dimp} to provide a probabilistic distribution prediction for the target’s location, which allows precise localization after occlusions and adapts to the various object scales. PrDiMP shows strong performance on challenges such as occlusions, out-of-view, and complex dynamic backgrounds, and achieves state-of-the-art results for tiny-sized objects.

\subsection{Impact of Fewer Large-scale Objects}
We further investigated the performance impact caused by a limited number of large-scale UAV objects, we evaluated trackers trained on CST Anti-UAV and those trained on Anti-UAV410 by testing them on the Anti-UAV410 test set, with the results shown in Figure~\ref{figure5}. Notably, our dataset achieved comparable success rates and accuracy scores with only 837 large-scale objects and 11,208 normal-sized objects, remarkably smaller training samples compared to Anti-UAV410's 196,735 large-scale objects and 97,942 normal-sized objects. This demonstrates that fewer large, prominent objects are sufficient to achieve good performance and further validates the effectiveness of the CST Anti-UAV dataset.

\section{Conclusion}

In this paper, we introduce the CST Anti-UAV dataset, a large-scale TIR benchmark specifically designed for UAV tracking. It consists of 220 videos with over 240k annotated bounding boxes, captured across diverse environments, times of day, and weather conditions. Notably, CST Anti-UAV is the first UAV tracking dataset to incorporate fully manual frame-level attribute annotations, addressing a critical gap in current benchmarks. The principal contributions of the CST Anti-UAV dataset include its focus on tiny UAV targets which are underrepresented in existing datasets, and the incorporation of multifaceted tracking scenarios spanning urban and wilderness environments, ensuring robust generalization. To validate its effectiveness, we retrained 20 recent SOT methods on both the CST Anti-UAV dataset and the existing largest anti-UAV tracking dataset, then conducted extensive experiments. The results show that the performance of these methods degrades significantly on our proposed dataset, and even the SOTA method struggles with tiny and inconspicuous objects in complex and dynamic backgrounds, highlighting the importance of the CST Anti-UAV for improving current UAV tracking methods. Overall, we believe the CST Anti-UAV benchmark will inspire the development of more robust UAV tracking methods and accelerate the deployment of reliable vision-based anti-UAV systems in the real world.

\noindent\textbf{Acknowledgements.} This work was supported in part by the National Natural Science Foundation of China (62222206, U2441241, 62401243, and 62272209), and the Jiangxi Province Natural Science Foundation (20242BAB212003). 

{\small
	\bibliographystyle{ieeenat_fullname}
	\bibliography{main}

@String(PAMI = {IEEE Trans. Pattern Anal. Mach. Intell.})

@String(CVPR= {IEEE Conf. Comput. Vis. Pattern Recog.})

@String(ICCV= {Int. Conf. Comput. Vis.})

@String(ECCV= {Eur. Conf. Comput. Vis.})

@String(NIPS= {Adv. Neural Inform. Process. Syst.})

@String(TIP  = {IEEE Trans. Image Process.})

@String(TMM  = {IEEE Trans. Multimedia})

@String(ACMMM= {ACM Int. Conf. Multimedia})

@String(AAAI = {AAAI})

@String(CVPRW= {IEEE Conf. Comput. Vis. Pattern Recog. Worksh.})

@String(PAMI  = {IEEE TPAMI})

@String(CVPR  = {CVPR})

@String(ICCV  = {ICCV})

@String(ECCV  = {ECCV})

@String(NIPS  = {NeurIPS})

@String(TIP   = {IEEE TIP})

@String(TMM   =	{IEEE TMM})

@String(ACMMM = {ACM MM})

@String(CVPRW= {CVPRW})

@article{uav5,
	Author = {Zhang, Zhen and Huang, Lehao and Wang, Qingwang and Jiang, Linhuan and
	Qi, Yemao and Wang, Shunyuan and Shen, Tao and Tang, Bo-Hui and Gu,
	Yanfeng},
	Title = {UAV Hyperspectral Remote Sensing Image Classification: A Systematic
	Review},
	Journal = {JSTARS},
	Year = {2025},
	Volume = {18},
	Pages = {3099-3124},
	DOI = {10.1109/JSTARS.2024.3522318},
	ISSN = {1939-1404},
	EISSN = {2151-1535},
	ResearcherID-Numbers = {Gu, Yanfeng/F-7781-2015
	},
	ORCID-Numbers = {Shen, Tao/0000-0003-1273-7950
	Zhang, Zhen/0000-0002-2300-7112
	huang, lehao/0009-0008-4239-2088
	Jiang, Linhuan/0009-0006-7723-478X
	Gu, Yanfeng/0000-0003-1625-7989
	TANG, Bo-Hui/0000-0002-1918-5346
	Wang, Qingwang/0000-0001-5820-5357},
	Unique-ID = {WOS:001398675100010},
}

@article{uav4,
	Author = {Han, Xiaoling and Lin, Bin and Na, Zhenyu and Li, Bowen and Zhang,
	Chaoyue and Zhang, Ran},
	Title = {Spatial Crowdsourcing-Based Task Allocation for UAV-Assisted Maritime
	Data Collection},
	Journal = {TVT},
	Year = {2025},
	Volume = {74},
	Number = {2},
	Pages = {3375-3388},
	Month = {FEB},
	DOI = {10.1109/TVT.2024.3483890},
	ISSN = {0018-9545},
	EISSN = {1939-9359},
	Unique-ID = {WOS:001425471500045},
}

@ARTICLE{uav3,
	author={Kumar, V. D. Ambeth and Ramachandran, Venkatesan and Rashid, Mamoon and Javed, Abdul Rehman and Islam, Shayla and Al Hejaili, Abdullah},
	journal={Tsinghua Science and Technology}, 
	title={An Intelligent Traffic Monitoring System in Congested Regions with Prioritization for Emergency Vehicle Using UAV Networks}, 
	year={2025},
	volume={30},
	number={4},
	pages={1387-1400},
	doi={10.26599/TST.2023.9010078}}

@inproceedings{tuav,   title={An autonomous vision-based target tracking system for rotorcraft unmanned aerial vehicles},  url={http://dx.doi.org/10.1109/iros.2017.8205986},  DOI={10.1109/iros.2017.8205986},  booktitle={IROS},  author={Cheng, Hui and Lin, Lishan and Zheng, Zhuoqi and Guan, Yuwei and Liu, Zhongchang},  year={2017},  month={Sep},  language={en-US}  }

@ARTICLE{uav2,
	author={Shi, Xiufang and Yang, Chaoqun and Xie, Weige and Liang, Chao and Shi, Zhiguo and Chen, Jiming},
	journal={IEEE Communications Magazine}, 
	title={Anti-Drone System with Multiple Surveillance Technologies: Architecture, Implementation, and Challenges}, 
	year={2018},
	volume={56},
	number={4},
	pages={68-74},
	doi={10.1109/MCOM.2018.1700430}}

@article{otb100,  
	title={Object Tracking Benchmark}, 
	url={http://dx.doi.org/10.1109/tpami.2014.2388226}, 
	DOI={10.1109/tpami.2014.2388226}, 
	journal={PAMI}, 
	author={Wu, Yi and Lim, Jongwoo and Yang, Ming-Hsuan}, 
	year={2015}, 
	month={Sep}, 
	pages={1834–1848}, 
	language={en-US} 
}

@inproceedings{otb50,
	   title={Online Object Tracking: A Benchmark},  url={http://dx.doi.org/10.1109/cvpr.2013.312},  DOI={10.1109/cvpr.2013.312},  
	   booktitle={CVPR},  
	   author={Wu, Yi and Lim, Jongwoo and Yang, Ming-Hsuan},  year={2013},  
	   month={Jun},  
	   language={en-US}  
	   }

@article{tc128,   title={Encoding Color Information for Visual Tracking: Algorithms and Benchmark},  url={http://dx.doi.org/10.1109/tip.2015.2482905},  DOI={10.1109/tip.2015.2482905},  journal={TIP},  author={Liang, Pengpeng and Blasch, Erik and Ling, Haibin},  year={2015},  month={Dec},  pages={5630–5644},  language={en-US}  }

@ARTICLE{alov,
	author={Smeulders, Arnold W. M. and Chu, Dung M. and Cucchiara, Rita and Calderara, Simone and Dehghan, Afshin and Shah, Mubarak},
	journal={PAMI}, 
	title={Visual Tracking: An Experimental Survey}, 
	year={2014},
	volume={36},
	number={7},
	pages={1442-1468},
	doi={10.1109/TPAMI.2013.230}}

@article{ytbb,   title={YouTube-BoundingBoxes: A Large High-Precision Human-Annotated Data Set for Object Detection in Video},  journal={Cornell University - arXiv},  author={Real, Esteban and Shlens, Jonathon and Mazzocchi, Stefano and Pan, Xin and Vanhoucke, Vincent},  year={2017},  month={Feb},  language={en-US}  }

@INPROCEEDINGS{lasot,
	author={Fan, Heng and Lin, Liting and Yang, Fan and Chu, Peng and Deng, Ge and Yu, Sijia and Bai, Hexin and Xu, Yong and Liao, Chunyuan and Ling, Haibin},
	booktitle={CVPR}, 
	title={LaSOT: A High-Quality Benchmark for Large-Scale Single Object Tracking}, 
	year={2019},
	volume={},
	number={},
	pages={5369-5378},
	doi={10.1109/CVPR.2019.00552}}

@ARTICLE{got10k,
	author={Huang, Lianghua and Zhao, Xin and Huang, Kaiqi},
	journal={PAMI}, 
	title={GOT-10k: A Large High-Diversity Benchmark for Generic Object Tracking in the Wild}, 
	year={2021},
	volume={43},
	number={5},
	pages={1562-1577},
	doi={10.1109/TPAMI.2019.2957464}}

@inproceedings{tracking,   title={TrackingNet: A Large-Scale Dataset and Benchmark for Object Tracking in the Wild},  url={http://dx.doi.org/10.1007/978-3-030-01246-5_19},  DOI={10.1007/978-3-030-01246-5_19},  booktitle={ECCV},  author={Müller, Matthias and Bibi, Adel and Giancola, Silvio and Alsubaihi, Salman and Ghanem, Bernard},  year={2018},  month={Jan},  pages={310–327},  language={en-US}  }

@article{ptb-tir,   title={PTB-TIR: A Thermal Infrared Pedestrian Tracking Benchmark},  url={http://dx.doi.org/10.1109/tmm.2019.2932615},  DOI={10.1109/tmm.2019.2932615},  journal={TMM},  author={Liu, Qiao and He, Zhenyu and Li, Xin and Zheng, Yuan},  year={2020},  month={Mar},  pages={666–675},  language={en-US}  }

@inproceedings{lsotb-tir,   title={LSOTB-TIR: A Large-Scale High-Diversity Thermal Infrared Object Tracking Benchmark},  url={http://dx.doi.org/10.1145/3394171.3413922},  DOI={10.1145/3394171.3413922},  booktitle={ACMMM},  author={Liu, Qiao and Li, Xin and He, Zhenyu and Li, Chenglong and Li, Jun and Zhou, Zikun and Yuan, Di and Li, Jing and Yang, Kai and Fan, Nana and Zheng, Feng},  year={2020},  month={Oct},  language={en-US}  }

@inproceedings{vottir2016,   title={The Thermal Infrared Visual Object Tracking VOT-TIR2016 Challenge Results},  url={http://dx.doi.org/10.1007/978-3-319-48881-3_55},  DOI={10.1007/978-3-319-48881-3_55},  booktitle={ECCVW},  author={Felsberg, Michael and Kristan, Matej and Matas, Jiři and Leonardis, Aleš and Pflugfelder, Roman and Häger, Gustav and Berg, Amanda and Eldesokey, Abdelrahman and Ahlberg, et al.},  year={2016},  month={Jan},  pages={824–849},  language={en-US}  }

@inproceedings{vottir2015,   title={The Thermal Infrared Visual Object Tracking VOT-TIR2015 Challenge Results},  url={http://dx.doi.org/10.1109/iccvw.2015.86},  DOI={10.1109/iccvw.2015.86},  booktitle={ICCVW},  author={Felsberg, Michael and Berg, Amanda and Hager, Gustav and Ahlberg, Jorgen and Kristan, Matej and Matas, Jiri and Leonardis, Ales and Cehovin, Luka and Fernandez, Gustavo and Vojır, Tomas and Nebehay, et al.},  year={2015},  month={Dec},  language={en-US}  }

@inproceedings{pdt-atv,   title={People detection and tracking from aerial thermal views},  url={http://dx.doi.org/10.1109/icra.2014.6907094},  DOI={10.1109/icra.2014.6907094},  booktitle={ICRA},  author={Portmann, Jan and Lynen, Simon and Chli, Margarita and Siegwart, Roland},  year={2014},  month={May},  language={en-US}  }

@inproceedings{bu-tiv,   title={A Thermal Infrared Video Benchmark for Visual Analysis},  url={http://dx.doi.org/10.1109/cvprw.2014.39},  DOI={10.1109/cvprw.2014.39},  booktitle={CVPRW},  author={Wu, Zheng and Fuller, Nathan and Theriault, Diane and Betke, Margrit},  year={2014},  month={Jun},  language={en-US}  }

@ARTICLE{antiuav,
	author={Jiang, Nan and Wang, Kuiran and Peng, Xiaoke and Yu, Xuehui and Wang, Qiang and Xing, Junliang and Li, Guorong and Guo, Guodong and Ye, Qixiang and Jiao, Jianbin and Zhao, Jian and Han, Zhenjun},
	journal={TMM}, 
	title={Anti-UAV: A Large-Scale Benchmark for Vision-Based UAV Tracking}, 
	year={2023},
	volume={25},
	number={},
	pages={486-500},
	doi={10.1109/TMM.2021.3128047}}

@article{antiuav410,
	title={Anti-UAV410: A Thermal Infrared Benchmark and Customized Scheme for Tracking Drones in the Wild},
	author={Huang, Bo and Li, Jianan and Chen, Junjie and Wang, Gang and Zhao, Jian and Xu, Tingfa},
	journal={PAMI},
	year={2023},
	publisher={IEEE}
}

@inproceedings{transt,   title={Transformer Tracking},  url={http://dx.doi.org/10.1109/cvpr46437.2021.00803},  DOI={10.1109/cvpr46437.2021.00803},  booktitle={CVPR},  author={Chen, Xin and Yan, Bin and Zhu, Jiawen and Wang, Dong and Yang, Xiaoyun and Lu, Huchuan},  year={2021},  month={Jun},  language={en-US}  }

@inproceedings{hift,   title={HiFT: Hierarchical Feature Transformer for Aerial Tracking},  url={http://dx.doi.org/10.1109/iccv48922.2021.01517},  DOI={10.1109/iccv48922.2021.01517},  booktitle={ICCV},  author={Cao, Ziang and Fu, Changhong and Ye, Junjie and Li, Bowen and Li, Yiming},  year={2021},  month={Oct},  language={en-US}  }

@inproceedings{stark,   title={Learning Spatio-Temporal Transformer for Visual Tracking},  url={http://dx.doi.org/10.1109/iccv48922.2021.01028},  DOI={10.1109/iccv48922.2021.01028},  booktitle={ICCV},  author={Yan, Bin and Peng, Houwen and Fu, Jianlong and Wang, Dong and Lu, Huchuan},  year={2021},  month={Oct},  language={en-US}  }

@inproceedings{stmtrack,
	title={STMTrack: Template-free Visual Tracking with Space-time Memory Networks},
	author={Fu, Zhihong and Liu, Qingjie and Fu, Zehua and Wang, Yunhong},
	booktitle={CVPR},
	pages={13774--13783},
	year={2021}
}

@InProceedings{ocean,
	author = {Zhipeng Zhang and Houwen Peng and Jianlong Fu and Bing Li and Weiming Hu},
	title = {Ocean: Object-aware Anchor-free Tracking},
	booktitle = {ECCV},
	month = {August},
	year = {2020}
}

@inproceedings{siamban,
	title={Siamese Box Adaptive Network for Visual Tracking},
	author={Chen, Zedu and Zhong, Bineng and Li, Guorong and Zhang, Shengping and Ji, Rongrong},
	booktitle={CVPR},
	pages={6668--6677},
	year={2020}
}

@InProceedings{cgacd,
	author = {Du, Fei and Liu, Peng and Zhao, Wei and Tang, Xianglong},
	title = {Correlation-Guided Attention for Corner Detection Based Visual Tracking},
	booktitle = {CVPR},
	month = {June},
	year = {2020}
}

@article{swintrack,  
	title={SwinTrack: A Simple and Strong Baseline for Transformer Tracking}, 
	author={Lin, Liting and Fan, Heng and Xu, Yong and Ling, Haibin}, 
	language={en-US} ,
	journal={NIPS},
	year = {2022}
}

@article{cswintt,  
	title={Transformer Tracking with Cyclic Shifting Window Attention}, 
	author={Song, Zikai and Yu, Junqing and Chen, Yi-PingPhoebe and Yang, Wei}, 
	language={en-US} ,
	journal={CVPR},
	year = {2022}
}

@inproceedings{siamattn,   title={Deformable Siamese Attention Networks for Visual Object Tracking},  url={http://dx.doi.org/10.1109/cvpr42600.2020.00676},  DOI={10.1109/cvpr42600.2020.00676},  booktitle={CVPR},  author={Yu, Yuechen and Xiong, Yilei and Huang, Weilin and Scott, Matthew R.},  year={2020},  month={Jun},  language={en-US}  }

@inproceedings{dimp,
  title={Learning discriminative model prediction for tracking},
  author={Bhat, Goutam and Danelljan, Martin and Gool, Luc Van and Timofte, Radu},
  booktitle={ICCV},
  pages={6182--6191},
  year={2019}
}

@inproceedings{sbt,
	title={Correlation-aware deep tracking},
	author={Xie, Fei and Wang, Chunyu and Wang, Guangting and Cao, Yue and Yang, Wankou and Zeng, Wenjun},
	booktitle={CVPR},
	year={2022}
}

@article{supersbt,
	title={Correlation-Embedded Transformer Tracking: A Single-Branch Framework},
	author={Xie, Fei and Yang, Wankou and Wang, Chunyu and Chu, Lei and Cao, Yue and Ma, Chao and Zeng, Wenjun},
	journal={PAMI},
	year={2024}
}

@inproceedings{atom,  
	title={ATOM: Accurate Tracking by Overlap Maximization}, 
	url={http://dx.doi.org/10.1109/cvpr.2019.00479}, 
	DOI={10.1109/cvpr.2019.00479}, 
	booktitle={CVPR}, 
	author={Danelljan, Martin and Bhat, Goutam and Khan, Fahad Shahbaz and Felsberg, Michael}, 
	year={2019}, 
	month={Jun}, 
	language={en-US} 
}

@inproceedings{eco,   title={ECO: Efficient Convolution Operators for Tracking},  url={http://dx.doi.org/10.1109/cvpr.2017.733},  DOI={10.1109/cvpr.2017.733},  booktitle={CVPR},  author={Danelljan, Martin and Bhat, Goutam and Khan, Fahad Shahbaz and Felsberg, Michael},  year={2017},  month={Jul},  language={en-US}  }

@inproceedings{siamrpnpp,   title={SiamRPN++: Evolution of Siamese Visual Tracking With Very Deep Networks},  url={http://dx.doi.org/10.1109/cvpr.2019.00441},  DOI={10.1109/cvpr.2019.00441},  booktitle={CVPR},  author={Li, Bo and Wu, Wei and Wang, Qiang and Zhang, Fangyi and Xing, Junliang and Yan, Junjie},  year={2019},  month={Jun},  language={en-US}  }

@inproceedings{metasdnet,   title={Meta-Tracker: Fast and Robust Online Adaptation for Visual Object Trackers},  url={http://dx.doi.org/10.1007/978-3-030-01219-9_35},  DOI={10.1007/978-3-030-01219-9_35},  booktitle={ECCV},  author={Park, Eunbyung and Berg, Alexander C.},  year={2018},  month={Jan},  pages={587–604},  language={en-US}  }

@article{simtrack,
	title={Backbone is All Your Need: A Simplified Architecture for Visual Object Tracking},
	author={Chen, Boyu and Li, Peixia and Bai, Lei and Qiao, Lei and Shen, Qiuhong and Li, Bo and Gan, Weihao and Wu, Wei and Ouyang, Wanli},
	journal={ECCV},
	year={2022}
}

@inproceedings{mdnet,   title={Learning Multi-Domain Convolutional Neural Networks for Visual Tracking},  url={http://dx.doi.org/10.1109/cvpr.2016.465},  DOI={10.1109/cvpr.2016.465},  booktitle={CVPR},  author={Nam, Hyeonseob and Han, Bohyung},  year={2016},  month={Jun},  language={en-US}  }

@inproceedings{udt,   title={Unsupervised Deep Tracking},  url={http://dx.doi.org/10.1109/cvpr.2019.00140},  DOI={10.1109/cvpr.2019.00140},  booktitle={CVPR},  author={Wang, Ning and Song, Yibing and Ma, Chao and Zhou, Wengang and Liu, Wei and Li, Houqiang},  year={2019},  month={Jun},  language={en-US}  }

@inproceedings{kys,
	author    = {Goutam Bhat and
	Martin Danelljan and
	Luc Van Gool and
	Radu Timofte},
	title     = {Know Your Surroundings: Exploiting Scene Information for Object Tracking},
	booktitle   = {ECCV},
	year      = {2020},
	url       = {https://arxiv.org/abs/2003.11014},
}

@inproceedings{keeptrack,  
	title={Learning Target Candidate Association to Keep Track of What Not to Track}, 
	url={http://dx.doi.org/10.1109/iccv48922.2021.01319}, 
	DOI={10.1109/iccv48922.2021.01319}, 
	booktitle={ICCV}, 
	author={Mayer, Christoph and Danelljan, Martin and Pani Paudel, Danda and Van Gool, Luc}, 
	year={2021}, 
	month={Oct}, 
	language={en-US} 
}

@article{globaltrack,   title={GlobalTrack: A Simple and Strong Baseline for Long-term Tracking},  url={http://dx.doi.org/10.1609/aaai.v34i07.6758},  DOI={10.1609/aaai.v34i07.6758},  journal={AAAI},  author={Huang, Lianghua and Zhao, Xin and Huang, Kaiqi},  year={2020},  month={Jun},  pages={11037–11044},  language={en-US}  }

@article{dutantiuav,   title={DUT-AntiUAV},  author={Zhao, Jie and Zhang, Jingshu and Li, Dongdong and Wang, Dong}, journal={TITS} ,year={2022},  month={May},  language={en-US}  }

@inproceedings{ostrack,
	title={Joint Feature Learning and Relation Modeling for Tracking: A One-Stream Framework},
	author={Ye, Botao and Chang, Hong and Ma, Bingpeng and Shan, Shiguang and Chen, Xilin},
	booktitle={ECCV},
	year={2022}
}

@inproceedings{aiatrack,
	title={AiATrack: Attention in Attention for Transformer Visual Tracking},
	author={Gao, Shenyuan and Zhou, Chunluan and Ma, Chao and Wang, Xinggang and Yuan, Junsong},
	booktitle={ECCV},
	pages={146--164},
	year={2022},
	organization={Springer}
}

@INPROCEEDINGS{mixformer,
	author={Cui, Yutao and Jiang, Cheng and Wang, Limin and Wu, Gangshan},
	booktitle={CVPR}, 
	title={MixFormer: End-to-End Tracking with Iterative Mixed Attention}, 
	year={2022},
	volume={},
	number={},
	pages={13598-13608},
	doi={10.1109/CVPR52688.2022.01324}}

@INPROCEEDINGS{mixformerv2,  booktitle={NIPS},  title={MixFormerV2: Efficient Fully Transformer Tracking},  author={Cui, Yutao and Song, Tianhui and Wu, Gangshan and Wang, Limin},  language={en-US} ,	year={2023}, }

@inproceedings{droptrack,
	title={DropMAE: Masked Autoencoders with Spatial-Attention Dropout for Tracking Tasks},
	author={Qiangqiang Wu and Tianyu Yang and Ziquan Liu and Baoyuan Wu and Ying Shan and Antoni B. Chan},
	booktitle={CVPR},
	year={2023}
}

@InProceedings{artrack,
	author    = {Wei, Xing and Bai, Yifan and Zheng, Yongchao and Shi, Dahu and Gong, Yihong},
	title     = {Autoregressive Visual Tracking},
	booktitle = {CVPR},
	month     = {June},
	year      = {2023},
	pages     = {9697-9706}
}

@InProceedings{seqtrack,
	title={SeqTrack: Sequence to Sequence Learning for Visual Object Tracking},
	author={Chen, Xin and Peng, Houwen and Wang, Dong and Lu, Huchuan and Hu, Han},
	booktitle={CVPR},
	year={2023}
}

@inproceedings{grm,
	title={Generalized Relation Modeling for Transformer Tracking},
	author={Gao, Shenyuan and Zhou, Chunluan and Zhang, Jun},
	booktitle={CVPR},
	pages={18686--18695},
	year={2023}
}

@InProceedings{romtrack,
	author    = {Cai, Yidong and Liu, Jie and Tang, Jie and Wu, Gangshan},
	title     = {Robust Object Modeling for Visual Tracking},
	booktitle = {ICCV},
	month     = {October},
	year      = {2023},
	pages     = {9589-9600}
}

@inproceedings{hiptrack,
	title={HIPTrack: Visual Tracking with Historical Prompts},
	author={Cai, Wenrui and Liu, Qingjie and Wang, Yunhong},
	booktitle={CVPR},
	year={2024}
}

@inproceedings{aqatrack,
	title={Autoregressive Queries for Adaptive Tracking with Spatio-Temporal Transformers},
	author={Xie, Jinxia and Zhong, Bineng and Mo, Zhiyi and Zhang, Shengping and Shi, Liangtao and Song, Shuxiang and Ji, Rongrong},
	booktitle={CVPR},
	pages={19300--19309},
	year={2024}
}

@InProceedings{artrackv2,
	author    = {Bai, Yifan and Zhao, Zeyang and Gong, Yihong and Wei, Xing},
	title     = {ARTrackV2: Prompting Autoregressive Tracker Where to Look and How to Describe},
	booktitle = {CVPR},
	month     = {June},
	year      = {2024}
}

@inproceedings{alpha,   title={Alpha-Refine: Boosting Tracking Performance by Precise Bounding Box Estimation},  url={http://dx.doi.org/10.1109/cvpr46437.2021.00525},  DOI={10.1109/cvpr46437.2021.00525},  booktitle={CVPR},  author={Yan, Bin and Zhang, Xinyu and Wang, Dong and Lu, Huchuan and Yang, Xiaoyun},  year={2021},  month={Jun},  language={en-US}  }

@inproceedings{siamfc,  
	title={Fully-Convolutional Siamese Networks for Object Tracking}, 
	url={http://dx.doi.org/10.1007/978-3-319-48881-3_56}, 
	DOI={10.1007/978-3-319-48881-3_56}, 
	booktitle={ECCV}, 
	author={Bertinetto, Luca and Valmadre, Jack and Henriques, João F. and Vedaldi, Andrea and Torr, Philip H. S.}, 
	year={2016}, 
	month={Jan}, 
	pages={850–865}, 
	language={en-US} 
}

@article{siamfc++,   title={SiamFC++: Towards Robust and Accurate Visual Tracking with Target Estimation Guidelines},  url={http://dx.doi.org/10.1609/aaai.v34i07.6944},  DOI={10.1609/aaai.v34i07.6944},  journal={AAAI},  author={Xu, Yinda and Wang, Zeyu and Li, Zuoxin and Yuan, Ye and Yu, Gang},  year={2020},  month={Jun},  pages={12549–12556},  language={en-US}  }

@inproceedings{prdimp,
  title={Probabilistic regression for visual tracking},
  author={Danelljan, Martin and Gool, Luc Van and Timofte, Radu},
  booktitle={CVPR},
  pages={7183--7192},
  year={2020}
}

@inproceedings{tomp,   title={Transforming Model Prediction for Tracking},  author={Mayer, Christoph and Danelljan, Martin and Bhat, Goutam and Paul, Matthieu and Pani, Danda and Fisher, Paudel and Luc, Yu and Gool, Van},  booktitle={CVPR},  year={2022} ,language={en-US}  }

@ARTICLE{refocus,
	author={Lai, Simiao and Liu, Chang and Wang, Dong and Lu, Huchuan},
	journal={TNNLS}, 
	title={Refocus the Attention for Parameter-Efficient Thermal Infrared Object Tracking}, 
	year={2024},
	volume={},
	number={},
	pages={1-12},
	doi={10.1109/TNNLS.2024.3420928}}
}

\end{document}